\documentclass{article}
\usepackage[final]{colm2026_conference}

\usepackage{microtype}
\usepackage{hyperref}
\usepackage{url}
\usepackage{booktabs}
\usepackage{graphicx}
\usepackage{makecell}
\usepackage{amsmath}
\usepackage{amsfonts}
\usepackage{algorithm}
\usepackage{algorithmic}
\usepackage{multirow} 

\usepackage{caption}

\usepackage{lineno}

\definecolor{darkblue}{rgb}{0, 0, 0.5}
\definecolor{ForestGreen}{RGB}{34,139,34}
\hypersetup{colorlinks=true, citecolor=darkblue, linkcolor=darkblue, urlcolor=darkblue}
\newcommand{\figdir}{figures}

\title{Unlocking Parallelism in Autoregressive Language Models via Speculative Decoding with Progressive Tree Drafting}

\author{
\begin{tabular}[t]{l}
Zipeng Gao\textsuperscript{1}
\quad Zhi Zheng\textsuperscript{1}\thanks{Corresponding author}
\quad Qingrong Xia\textsuperscript{2}
\quad Junda Lin\textsuperscript{1}
\quad Ziwei Zhao\textsuperscript{1}
\\[4pt]
Tong Xu\textsuperscript{1}
\quad Zhefeng Wang\textsuperscript{2}
\quad Enhong Chen\textsuperscript{1}
\\[8pt]
\textsuperscript{1}State Key Laboratory of Cognitive Intelligence, \\
University of Science and Technology of China
\\[2pt]
\textsuperscript{2}Huawei Technologies Co., Ltd.
\\[4pt]
{\normalfont\small\texttt{\{gaozp619,zhengzhi97,linjunda,zzw22222\}@mail.ustc.edu.cn}}
\\[2pt]
{\normalfont\small\texttt{\{zhengzhi97,tongxu,cheneh\}@ustc.edu.cn}}
\\[2pt]
{\normalfont\small\texttt{\{xiaqingrong,wangzhefeng\}@huawei.com}}
\\[2pt]
\end{tabular}
}

\begin{document}

\ifcolmsubmission
\linenumbers
\fi

\maketitle

\begin{abstract}
Speculative decoding has significantly accelerated Large Language Model (LLM) inference by alleviating memory-bound bottlenecks. However, traditional speculative decoding typically relies on auxiliary draft modules, incurring significant training and communication overhead. Although recent methods attempt to generate drafts within the target model itself, they often fail to fully exploit its latent parallel capacity due to a lack of structural coordination. In this paper, we propose \textbf{Progressive Tree Drafting (PTD)}, which employs a structured, guided parallel drafting strategy to harness the model's parallel potential. By coupling a progressive tree structure with a stepwise pruning mechanism, PTD actively guides the LLM to explore multiple semantic paths in a single forward pass, ensuring both draft diversity and coherence. Experiments demonstrate that PTD achieves up to $2\times$ decoding speedup across various benchmarks while remaining training-free and model-agnostic. Our code is available at: https://github.com/MINE-USTC/PTD.

\end{abstract}

\section{Introduction}
Speculative decoding has emerged as an important paradigm for accelerating Large Language Model (LLM) inference, mitigating the memory-bound bottlenecks inherent in autoregressive generation~\citep{yuan2024llm,xia-etal-2024-unlocking}. By transforming the inefficient token-by-token processing into a candidate parallel verification process, this paradigm enhances computational utility and inference speed without compromising generation quality.

The key to speculative decoding is obtaining high-quality drafts. Traditional approaches typically employ smaller auxiliary modules to generate drafts, which often incur significant communication overhead and require substantial training effort for model alignment~\citep{xia-etal-2023-speculative,leviathan2023fast,chen2023accelerating,Miao_2024,yang2024multi,cai2024medusa,stern2018blockwise,li2025eaglespeculativesamplingrequires}. 
To mitigate these issues, recent research has shifted toward training-free, model-agnostic strategies that generate drafts directly within the target LLM itself. 
For instance, Lookahead Decoding (LADE)~\citep{fu2024break} supplements the primary decoding objective with a parallel Jacobi iteration task to predict and refine linear candidate sequences, while Self-Draft~\citep{gao2025multi} leverages the intrinsic robustness of LLMs to incorporate an auxiliary drafting task, generating multiple candidate branches through input perturbations.

\begin{figure}[t]
    \centering
    \includegraphics[width=0.8\linewidth]{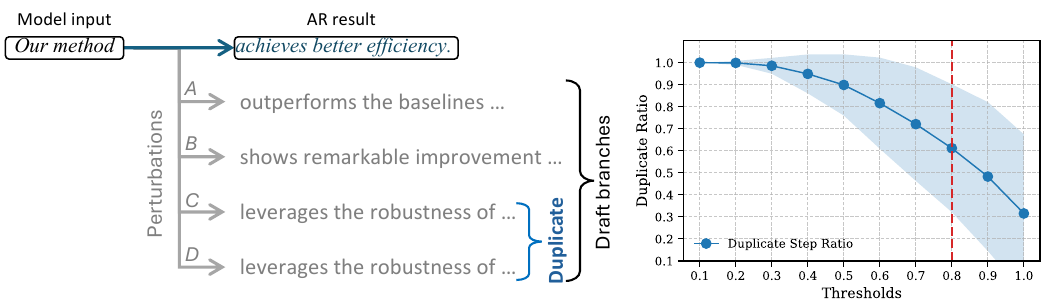}
    \caption{A drafting example (left) and the branch similarity analysis (right) of Self-Draft~\citep{gao2025multi}. The curve means the proportion of steps with at least two branches above the similarity threshold.}
    \label{fig:example-ana}
\end{figure}

While these methods confirm the feasibility of endogenous acceleration, we argue that the latent parallel processing capabilities inherent in AR models remain under-exploited, primarily due to the independent and unstructured nature of their candidate generation. Specifically, our in-depth analysis reveals a critical bottleneck in this paradigm. As quantified in Figure~\ref{fig:example-ana} (right), our analysis of Self-Draft shows that more than half of the decoding steps contain branches with over 80\% similarity. This high level of semantic redundancy suggests that simply generating independent linear branches fails to effectively direct the model's parallel resources toward diverse drafting. 
This observation motivates us to rethink: how can we more intentionally and structurally guide the model to transform its parallel potential into high-quality, diverse drafts?

\begin{figure}[t]
    \centering
    \includegraphics[width=0.8\linewidth]{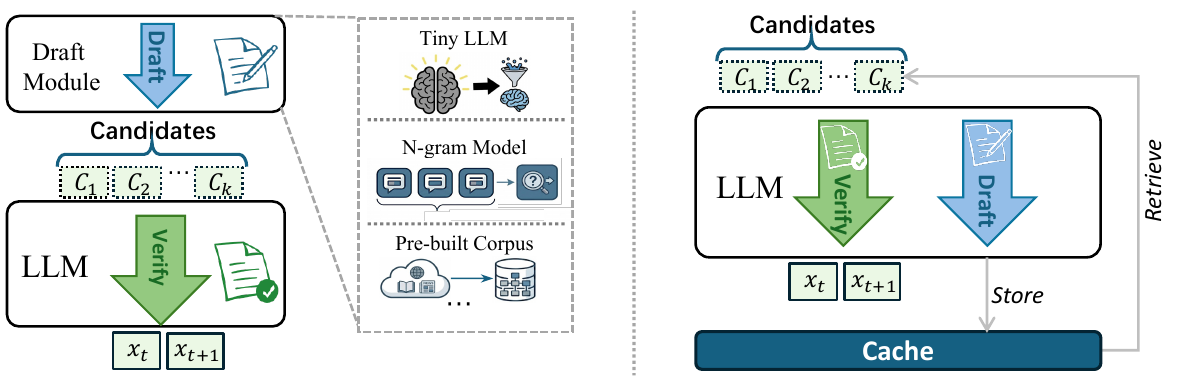}
    \caption{Paradigms of speculative decoding. (Left) Traditional methods using auxiliary draft modules. (Right) Endogenous, training-free methods.}
    \label{fig:method-compare}
\end{figure}

Building on this insight, we propose Progressive Tree Drafting (PTD), a novel strategy that reformulates drafting as a structured, guided parallel inference process. By leveraging a tree structure to merge redundant prefixes, PTD eliminates the computational waste inherent in independent branching. Furthermore, the integration of a progressively updated evolution with a stepwise pruning mechanism allows PTD to guide the LLM to explore multiple distinct semantic paths within a single forward pass. This approach ensures the generated drafts are both diverse and coherent, thereby increasing the acceptance rate and unlocking the model’s parallel potential.

The main contributions of this paper are summarized as follows:
\begin{itemize}
    \item We identify and quantify the computational redundancy in existing linear or unstructured drafting methods, revealing that they fail to fully exploit the parallel capacity of AR models.
    \item We introduce Progressive Tree Drafting (PTD), a strategy that reformulates drafting into a structured, guided parallel generation process to ensure both draft diversity and coherence.
    \item Experimental results demonstrate that PTD consistently achieves up to a $2\times$ speedup across various benchmarks without requiring any auxiliary modules.
\end{itemize}

The structure of this paper is as follows: First, we review related works, followed by a detailed description of the proposed method. Then, we present experimental results to validate its effectiveness. Finally, we conclude the paper and discuss potential directions for future research.

\section{Related Works}
\label{gen_inst}

\subsection{Speculative Decoding with Additional Modules}

The conventional speculative decoding paradigm relies on \textbf{additional modules}, including independent models and architectural extensions, to generate candidate tokens.
The initial formulation~\citep{xia-etal-2023-speculative,leviathan2023fast} employs a smaller, independent \textit{draft model} to generate candidates, which are then verified by the target LLM. These methods suffer from serialization bottlenecks and heavy reliance on the draft model's accuracy. PEARL~\citep{liu2025pearlparallelspeculativedecoding} and
SwiftSpec~\citep{zhang2025swiftspecultralowlatencyllm} addresses these synchronization issues by optimizing the interaction protocol between the drafting and verification stages.
Subsequent works like MCSD~\citep{yang2024multi} and SpecInfer~\citep{Miao_2024} extend this by using multiple draft models or sampling strategies to boost diversity, yet the maintenance of separate models remains a bottleneck. 
JudgeDecoding~\citep{bachmann2025judgedecodingfasterspeculative} adopts a relaxed alignment paradigm to boost verification speed via a marginal accuracy loss.
Retrieval-based methods such as REST~\citep{he2023rest} substitute the draft model with an external datastore module, but this shifts the dependency to the quality and domain relevance of the retrieved corpus. 

Alternatively, some approaches integrate the drafting module directly into the target model's architecture. The EAGLE series~\citep{10.5555/3692070.3693232,li2024eagle2fasterinferencelanguage}, Medusa~\citep{cai2024medusa}, Blockwise Decoding~\citep{stern2018blockwise}, Hydra~\citep{ankner2024hydrasequentiallydependentdraftheads} attach \textit{additional prediction heads or layers} to the LLM to predict future tokens in parallel. Although these architectural modifications reduce communication latency compared to independent models, they fundamentally alter the model structure, requiring substantial extra training and preventing seamless deployment on off-the-shelf LLMs.

\subsection{Training-Free Endogenous Acceleration}
In contrast to module-based approaches, a recent line of research focuses on \textit{endogenous acceleration}, which exploits the target LLM's latent capabilities to generate drafts without any additional modules or training.
Lookahead Decoding (LADE)~\citep{fu2024break} utilizes the Jacobi iteration method to perform parallel decoding within the original model architecture. 
Parallelly, Self-Draft~\citep{gao2025multi} leverages the intrinsic robustness of LLMs, employing multi-branch input perturbations to induce the model to generate its own drafts. 

These methods represent a shift towards architecture-agnostic acceleration. However, as discussed in the Introduction, current endogenous methods often rely on linear or iterative generation schemes that lack structural coordination, failing to fully harness the model's parallel potential. Our Progressive Tree Drafting (PTD) advances this paradigm by introducing a structured, guided exploration mechanism, achieving efficient self-acceleration without the burden of extra modules.

\section{Methodology}
\label{headings}

Conventional autoregressive decoding predicts one token at a time. Given a prefix $\mathbf{X} = [x_1, x_2, \cdots, x_{t-1}]$, the LLM computes the next-token distribution $P(y_t|\mathbf{X})$ and selects the next token with a decoding strategy $\mathcal{S}$:
$$x_t = \mathcal{S}(P(y_t|\mathbf{X}))$$
Speculative decoding generalizes this step by introducing candidate continuations $\mathbf{C}_{\mathbf{X}}$ and verifying them in parallel:
$$
x_t, x_{t+1},...,x_{t+k} = \mathcal{S}\left(P(y_t, \mathbf{y}_{\mathbf{C}}|[\mathbf{X};\mathbf{C}_{\mathbf{X}}])\right),
$$

These formulations provide the theoretical foundation for speculative decoding. In practice, traditional implementations typically rely on external draft models, yet this dependency incurs alignment and communication overhead. Endogenous acceleration offers a more efficient alternative by leveraging the target model's internal parallel processing capacity.

Specifically, the inference behavior of Transformer-based LLMs is governed by the input content and the attention matrix. By reconfiguring the input content and the attention mask, we can transfer the model's attention from a linear path and to a multiple, concurrent reasoning trajectories. This flexibility allows the LLM to explore a structured semantic space and generate high-quality drafts in a single forward pass.

Based on this insight, we develop Progressive Tree Drafting (PTD) to transform the model's parallel potential into actual decoding speedup. The design of PTD focuses on solving two practical issues: how to organize the parallel drafting paths and how to maintain their contextual coherence. For the former, we adopt a drafting tree, as its prefix-sharing nature represents the most complex topology theoretically supported by autoregressive models. For the latter, we introduce a progressive update algorithm with stepwise pruning to ensure the generated content remains context-relevant. 

\begin{figure*}
    \centering
    \includegraphics[width=\linewidth]{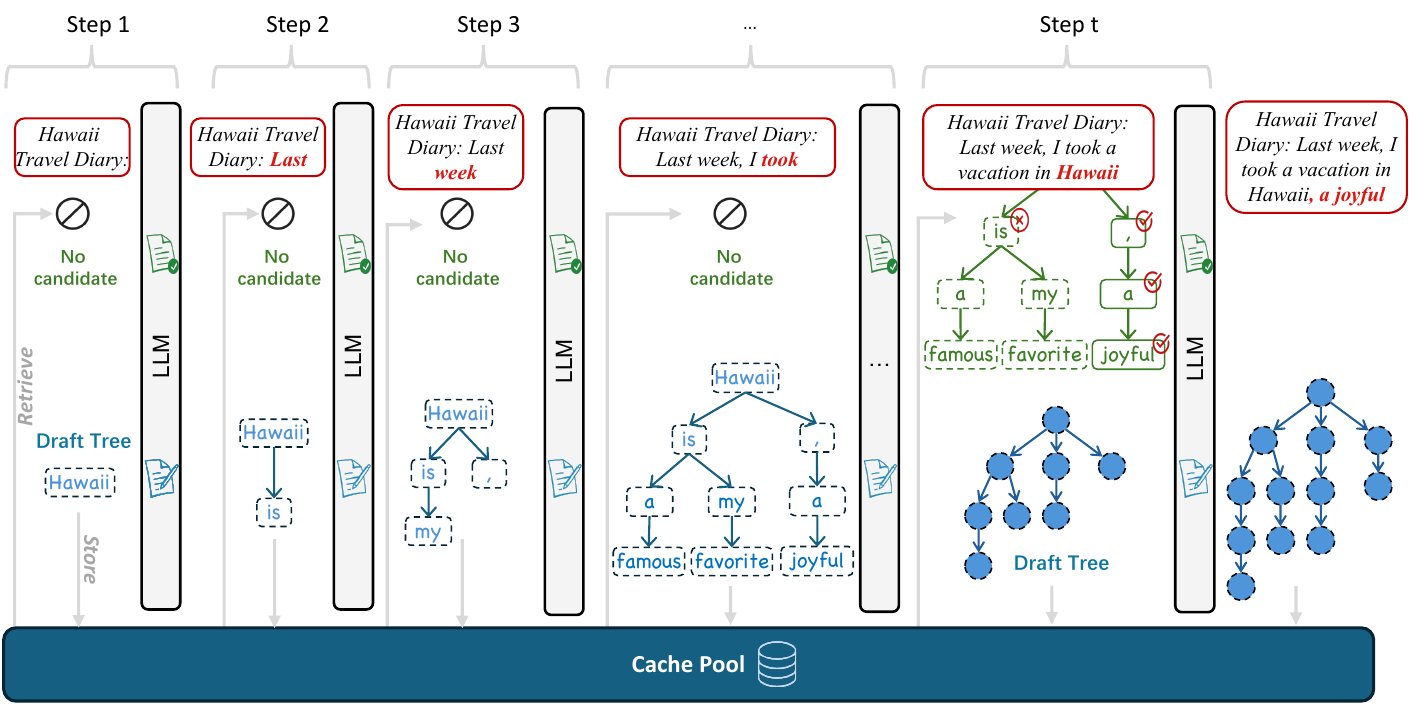}
    \caption{Overview of the Progressive Tree Drafting (PTD) framework. The \textcolor{ForestGreen}{green} pathway illustrates the verification process of candidate drafts, while the \textcolor{blue}{blue} pathway denotes the semantic-guided progressive tree drafting process. The \textcolor{red}{red} box indicates the decoded results.
   }
    \label{fig:OV}
\end{figure*}

Figure~\ref{fig:OV} illustrates the overall framework of PTD. In the following subsections, we begin by introducing the tree expansion and pruning strategies that regulate the dynamic evolution of the drafting tree. Then, we detail the draft extraction process and the verification mechanism.

\subsection{Progressive Tree Drafting}

To transform the parallel potential of LLMs into actual speedup, we propose the Progressive Tree Drafting (PTD) mechanism. This section details the tree's lifecycle, from its structural initialization to its dynamic evolution and pruning.

\textbf{Tree Construction.} We represent the drafting structure as a tree $T = (V, E)$. To bootstrap the progressive process, $T$ is first initialized with a set of seed nodes. These nodes can be generated stochastically or derived from semantic information in the context, serving to trigger the LLM's reasoning across diverse semantic trajectories.

To ensure semantic consistency, each node $v \in V$ must only attend to its ancestors. 
Formally, for any node $v$, the set of its observable prefix nodes $\pi(v)$ is defined recursively:
$$\pi(v) = \{ v \} \cup \pi(\mathcal{P}(v))$$
where $\mathcal{P}(v)$ denotes the unique parent of node $v$. The positional encoding for node $v$ is determined by the cardinality of $\pi(v)$, ensuring that the tree-structured input remains compatible with the Transformer's causal attention.

\textbf{Progressive Expansion.} The inference on the draft tree $T^{t-1}$ at step $t$ is formulated as:
$$x_t, \mathcal{D}_{t} = \mathcal{S}\left(P\left(y_t,\mathbf{y}_{T}|[\mathbf{X};T^{t-1}]\right)\right)$$
where $\mathcal{D}_{t} = \{d_v \mid v \in V^{t-1}\}$ is the set of \textbf{draft tokens} generated by all nodes in $T^{t-1}$ in a single forward pass. PTD then evolves the draft tree to $T^t = (V^t, E^t)$ by appending these newly generated tokens as child nodes:
$$V^t = V^{t-1} \cup \{d_v\}_{v \in V^{t-1}}, \quad E^t = E^{t-1} \cup \{(v, d_v)\}_{v \in V^{t-1}}$$

\begin{figure}[htbp]
    \centering
    \includegraphics[width=0.6\linewidth]{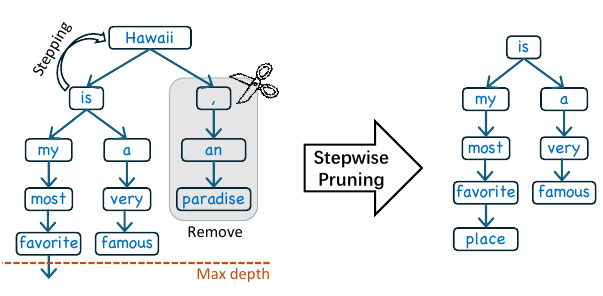}
    \caption{Illustration stepwise prune algorithms.}
    \label{fig:PDT}
\end{figure}

\textbf{Overhead Constraints.} Generally, the number of nodes in the draft tree ensures the diversity of the drafts it generates, and the expansion process maintains the semantic coherence between the adjacent nodes in the tree. However, the computational overhead introduced by the draft tree increases progressively as it grows. Hence, it is necessary to impose constraints on its growth to prevent excessive size, which could otherwise degrade the overall decoding speed. Specifically, we impose constraints on the tree's topology along two dimensions: 
\begin{itemize} 
    \item \textbf{Width Control:} We limit the max number of child nodes for each parent. This prevents low-confidence tokens from branching excessively, ensuring that the tree focuses on high-probability reasoning paths. 
    \item \textbf{Stepping Mechanism (Depth Control):} To prevent the tree from becoming overly deep, we implement a sliding-window-style stepping mechanism. As illustrated in Figure~\ref{fig:PDT}, when a sub-tree exceeds a depth threshold, we retain the earliest-added child and its descendants as the new sub-tree, pruning all other stale branches. This stepping-wise prune mechanism helps preserve the semantic coherence and contextual relevance of the draft tree.
\end{itemize}
These parameters allow for fine-grained control over the draft tree's complexity to suit different model scales and application requirements. In the following experimental section, we provide a thorough evaluation of how these constraints influence the trade-off between drafting efficiency and overall speedup.

\textbf{Draft Extraction via Merging.} Finally, we extract semantic subtrees from the expanded draft tree $T^t$ and aggregate them with the existing candidates in the draft cache pool. 
Specifically, any subtree $T'$ in the draft tree $T^{t}$ will be merged with the cached candidate tree that shares the same root node value. We define the following recursive merging function $\mathcal{M}$ for any two trees $T$ and $T'$ with same root $r$:
\[
        \mathcal{M}(T,T') = \begin{cases}
           \left(V \cup v, E \cup (r,v)\right), \forall v \in \sigma(T') \setminus\sigma(T)\\
           \mathcal{M}(T_v,T'_v), \forall v \in \sigma(T') \cap \sigma(T)
        \end{cases},
\]
where $\sigma(T)$ denotes the set of direct child nodes of the tree $T$, and $T_v$ is the subtree with root of $v$ in tree $T$.

\subsection{Candidate Verification}

In parallel with the autoregressive decoding process and the progressive tree drafting process, a candidate tree validation process is concurrently executed during the forward pass. 
Given the partially decoded token sequence $\mathbf{X}$, we retrieve corresponding drafts from the draft pool, forming the candidate tree $\mathbf{C}_{\mathbf{X}}$.

To verify this candidate tree, we apply the similar attention mask and positional encoding strategy as used in the drafting process. Consequently, after a forward pass through the LLM, each node in $\mathbf{C}_{\mathbf{X}}$ produces a verification token conditioned on its prefix. Together with the autoregressive decoding process and the progressive tree drafting process, we formulate the overall model forward process as follows:
$$
x_t, \mathcal{D}, \mathcal{V} = \mathcal{S}\left(P(y_t,\mathbf{y}_{T},\mathbf{y}_{\mathbf{C}}|[\mathbf{X};T^{i-1};\mathbf{C}_{\mathbf{X}}])\right),
$$
where $\mathcal{V}$ denotes the \textbf{verification tokens} that are generated by each node and its prefix in the candidate tree.

Finally, the accepted tokens $\mathbf{X}'$ can be obtained by identifying all eligible edges $\mathcal{I}$ in the candidate tree. Under the \textbf{greedy decoding strategy}, the verification tokens $\mathcal{V}$ for all nodes $V_{\mathbf{C}_\mathbf{X}}$ of the candidate tree $\mathbf{C}_\mathbf{X}$ are selected based on the model’s highest-probability predictions. 
Eligible edges are identified recursively by verifying whether a node’s verification token appears among its child nodes. That is: 
$$\mathcal{I} = \{(n,\mathcal{V}_n)|\mathcal{V}_n \in \sigma(n),\forall n \in V_{\mathbf{C}_\mathbf{X}}\}.$$

For the \textbf{sampling decoding strategy}, we determine whether each token is accepted using a without-replacement sampling method based on normalized probabilities, following an approach similar to LADE~\citep{fu2024break} and SpecInfer~\citep{Miao_2024}. Specifically, starting from the root node of the candidate draft tree, the LLM produces a probability distribution $P_v$ over the next token at each node $v$. Each node may have multiple successor nodes \( [c_1, c_2, ..., c_k] \), and a sampling process is iteratively applied to these \( k \) candidates.

At each iteration, a random number \( r \sim \mathcal{U}(0, 1) \) is drawn, and the candidates are traversed in order. If \( r \leq P_i \), the candidate \( c_i \) is selected, and the edge between \( c_i \) and its parent node is marked as eligible and appended to the eligible edge set $\mathcal{I}$. If not, \( P_i \) is set to zero, and the remaining probabilities are renormalized. This process continues until a candidate satisfies \( r \leq P_i \), ensuring that the final selection remains faithful to the original distribution. We provide the full PTD decoding algorithm, the recursive candidate-tree sampling algorithm, and the proof of distributional consistency in Appendices~\ref{alg-PTD},~\ref{alg-sample}, and~\ref{proof}.

The final accepted sequence is the path formed by eligible edges starting from the root node $n_{0}$. That is, 
$$\mathbf{X}' = (n_0,n_1,...n_k,\mathcal{V}_{n_k})$$
where $\forall i<k, (n_i, n_{i+1})\in \mathcal{I}$ and $\mathbf{X}'$ are the tokens we decoded in a single model forward pass.

\section{Experiments}

\subsection{Settings}

\textbf{Benchmarks.} To evaluate the performance of PTD across diverse scenarios, we utilize several representative benchmarks. For general conversation, we use MT-Bench~\citep{zheng2023judging}, which covers eight distinct task categories with 80 problems in total. For mathematical reasoning, we randomly sample 100 questions from the GSM-8k~\citep{cobbe2021gsm8k} dataset. For code generation, we employ the full HumanEval~\citep{chen2021codex} dataset and a 100-problem test subset from MBPP~\citep{austin2021program}.

\textbf{Baselines.} We compare PTD against several representative strategies. Standard Autoregressive (AR) decoding serves as the primary baseline, and the training-free baselines include Lookahead Decoding (LADE)~\citep{fu2024break} and Self-Draft~\citep{gao2025multi}. All baseline hyperparameters are kept at their default settings.

\textbf{Models.} 
For general conversation (MT-Bench) and mathematical reasoning (GSM-100), we employ LLaMA-2 (7B/13B), LLaMA-3 (8B), Qwen-2.5 (7B/14B/32B), and Qwen-3 (8B/14B). For code generation (HumanEval, MBPP-100), we utilize CodeLLaMA-7B/13B.

\textbf{Metrics.} 
We employ five metrics for evaluation: (i) Throughput (TP) (tokens/s) for end-to-end speed; (ii) Accept Length (AL) and (iii) Hit Rate (HR) to characterize draft coherence and diversity; (iv) Decoding Efficiency (DE), the average tokens produced per forward pass, defined as:
$$\text{DE} = \text{HR} \cdot \text{AL} + (1 - \text{HR})$$
and (v) Computational Overhead, the average additional tokens processed per step during drafting ($Dft$) and verification ($Ver$).

All experiments were conducted on NVIDIA L20 GPUs (48 GB RAM) using BF16 precision to enhance computational efficiency. Inference was performed consistently with a batch size of one throughout. Unless otherwise specified, all draft tokens are obtained via greedy (Top-1) and the draft tree is initialized randomly.
Additional analyses of sampling strategies and initialization are provided in Appendices~\ref{app:sample-strategy} and~\ref{app:init-strategy}.

\subsection{Results}

\subsubsection{Main Results}

\begin{table}[H]
    \centering
    \captionsetup{skip=3pt}
    \footnotesize
    \setlength{\tabcolsep}{4pt} 
    \renewcommand{\arraystretch}{1.30}
    \begin{tabular}{lcccccccc}
        \toprule
        \multicolumn{9}{c}{\textbf{MT-Bench}} \\
        \midrule
        & \textbf{L2-7B} & \textbf{L2-13B} & \textbf{L3-8B} & \textbf{Q2-7B} & \textbf{Q2-14B} & \textbf{Q2-32B} & \textbf{Q3-8B} & \textbf{Q3-14B} \\ \cmidrule{1-9}
        AR (Ref.) & \makecell{1.00$\times$ \\ \scriptsize(39\tiny$\pm$3.9)} & \makecell{1.00$\times$ \\ \scriptsize(24\tiny$\pm$1.7)} & \makecell{1.00$\times$ \\ \scriptsize(38\tiny$\pm$2.6)} & \makecell{1.00$\times$ \\ \scriptsize(36\tiny$\pm$4.3)} & \makecell{1.00$\times$ \\ \scriptsize(20\tiny$\pm$1.9)} & \makecell{1.00$\times$ \\ \scriptsize(10\tiny$\pm$0.6)} & \makecell{1.00$\times$ \\ \scriptsize(39\tiny$\pm$3.4)} & \makecell{1.00$\times$ \\ \scriptsize(24\tiny$\pm$1.3)} \\
        \cmidrule{1-9}
        LADE & \makecell{1.49$\times$ \\ \scriptsize(58\tiny$\pm$9.8)} & \makecell{1.38$\times$ \\ \scriptsize(33\tiny$\pm$4.9)} & \makecell{1.42$\times$ \\ \scriptsize(54\tiny$\pm$5.2)} & \makecell{1.42$\times$ \\ \scriptsize(51\tiny$\pm$8.8)} & \makecell{1.45$\times$ \\ \scriptsize(29\tiny$\pm$5.0)} & \makecell{1.60$\times$ \\ \scriptsize(16\tiny$\pm$3.0)} & \makecell{1.51$\times$ \\ \scriptsize(59\tiny$\pm$7.3)} & \makecell{1.29$\times$ \\ \scriptsize(31\tiny$\pm$4.7)} \\
        \cmidrule{1-9}
        Self-Draft & \makecell{1.54$\times$ \\ \scriptsize(60\tiny$\pm$12.1)} & \makecell{1.54$\times$ \\ \scriptsize(37\tiny$\pm$6.6)} & \makecell{1.53$\times$ \\ \scriptsize(58\tiny$\pm$7.9)} & \makecell{1.47$\times$ \\ \scriptsize(53\tiny$\pm$12.5)} & \makecell{1.50$\times$ \\ \scriptsize(30\tiny$\pm$6.6)} & \makecell{1.60$\times$ \\ \scriptsize(16\tiny$\pm$3.3)} & \makecell{1.54$\times$ \\ \scriptsize(60\tiny$\pm$9.5)} & \makecell{1.38$\times$ \\ \scriptsize(33\tiny$\pm$5.1)} \\
        \cmidrule{1-9}
        \textbf{PTD (Ours)} & \makecell{\textbf{1.67$\times$} \\ \scriptsize(\textbf{65}\tiny$\pm$\textbf{11.1})} & \makecell{\textbf{1.67$\times$} \\ \scriptsize(\textbf{40}\tiny$\pm$\textbf{5.8})} & \makecell{\textbf{1.66$\times$} \\ \scriptsize(\textbf{63}\tiny$\pm$\textbf{7.1})} & \makecell{\textbf{1.69$\times$} \\ \scriptsize(\textbf{61}\tiny$\pm$\textbf{15.4})} & \makecell{\textbf{1.75$\times$} \\ \scriptsize(\textbf{35}\tiny$\pm$\textbf{7.6})} & \makecell{\textbf{1.90$\times$} \\ \scriptsize(\textbf{19}\tiny$\pm$\textbf{3.9})} & \makecell{\textbf{1.64$\times$} \\ \scriptsize(\textbf{64}\tiny$\pm$\textbf{6.9})} & \makecell{\textbf{1.54$\times$} \\ \scriptsize(\textbf{37}\tiny$\pm$\textbf{6.2})} \\
        \midrule
        \multicolumn{9}{c}{\textbf{GSM-100}} \\
        \midrule
        & \textbf{L2-7B} & \textbf{L2-13B} & \textbf{L3-8B} & \textbf{Q2-7B} & \textbf{Q2-14B} & \textbf{Q2-32B} & \textbf{Q3-8B} & \textbf{Q3-14B} \\ 
        \cmidrule{1-9}
        AR (Ref.) & \makecell{1.00$\times$ \\ \scriptsize(43\tiny$\pm$0.9)} & \makecell{1.00$\times$ \\ \scriptsize(26\tiny$\pm$0.4)} & \makecell{1.00$\times$ \\ \scriptsize(41\tiny$\pm$0.6)} & \makecell{1.00$\times$ \\ \scriptsize(39\tiny$\pm$1.9)} & \makecell{1.00$\times$ \\ \scriptsize(22\tiny$\pm$0.5)} & \makecell{1.00$\times$ \\ \scriptsize(10\tiny$\pm$0.2)} & \makecell{1.00$\times$ \\ \scriptsize(40\tiny$\pm$1.3)} & \makecell{1.00$\times$ \\ \scriptsize(25\tiny$\pm$0.6)} \\
        \cmidrule{1-9}
        LADE & \makecell{1.70$\times$ \\ \scriptsize(73\tiny$\pm$5.6)} & \makecell{1.58$\times$ \\ \scriptsize(41\tiny$\pm$3.3)} & \makecell{1.71$\times$ \\ \scriptsize(70\tiny$\pm$5.2)} & \makecell{1.56$\times$ \\ \scriptsize(61\tiny$\pm$6.3)} & \makecell{1.55$\times$ \\ \scriptsize(34\tiny$\pm$3.8)} & \makecell{1.80$\times$ \\ \scriptsize(18\tiny$\pm$1.5)} & \makecell{1.73$\times$ \\ \scriptsize(69\tiny$\pm$5.9)} & \makecell{1.44$\times$ \\ \scriptsize(36\tiny$\pm$2.7)} \\
        \cmidrule{1-9}
        Self-Draft & \makecell{1.72$\times$ \\ \scriptsize(74\tiny$\pm$6.2)} & \makecell{1.69$\times$ \\ \scriptsize(44\tiny$\pm$4.5)} & \makecell{1.80$\times$ \\ \scriptsize(74\tiny$\pm$5.0)} & \makecell{1.56$\times$ \\ \scriptsize(61\tiny$\pm$7.4)} & \makecell{1.59$\times$ \\ \scriptsize(35\tiny$\pm$4.2)} & \makecell{1.90$\times$ \\ \scriptsize(19\tiny$\pm$1.6)} & \makecell{1.83$\times$ \\ \scriptsize(73\tiny$\pm$5.7)} & \makecell{1.52$\times$ \\ \scriptsize(38\tiny$\pm$3.8)} \\
        \cmidrule{1-9}
        \textbf{PTD (Ours)} & \makecell{\textbf{1.91$\times$} \\ \scriptsize(\textbf{82}\tiny$\pm$\textbf{7.5})} & \makecell{\textbf{1.85$\times$} \\ \scriptsize(\textbf{48}\tiny$\pm$\textbf{4.2})} & 
        \makecell{\textbf{1.93$\times$} \\ \scriptsize(\textbf{79}\tiny$\pm$\textbf{6.7})} & \makecell{\textbf{1.87$\times$} \\ \scriptsize(\textbf{73}\tiny$\pm$\textbf{13.3})} & 
        \makecell{\textbf{1.86$\times$} \\ \scriptsize(\textbf{41}\tiny$\pm$\textbf{5.9})} & \makecell{\textbf{2.30$\times$} \\ \scriptsize(\textbf{23}\tiny$\pm$\textbf{2.1})} & 
        \makecell{\textbf{1.93$\times$} \\ \scriptsize(\textbf{77}\tiny$\pm$\textbf{9.4})} & \makecell{\textbf{1.76$\times$} \\ \scriptsize(\textbf{44}\tiny$\pm$\textbf{5.2})} \\
        \bottomrule
    \end{tabular}
    \caption{Speedup performance on MT-Bench and GSM-100 across Llama and Qwen model series. The primary metric is the Speedup Ratio ($n\times$), with throughput (tokens/s) and standard deviation shown in small parentheses.}
    \label{tab:chat-math-res}
\end{table}

\begin{table}[H]
    \centering
    \captionsetup{skip=3pt}
    \footnotesize 
    \setlength{\tabcolsep}{1.5pt}
    \renewcommand{\arraystretch}{1.15}
    \begin{tabular}{lcccc}
        \toprule
        \multirow{2}{*}{\textbf{Method}} & \multicolumn{2}{c}{\textbf{HumanEval}} & \multicolumn{2}{c}{\textbf{MBPP-100}} \\
        \cmidrule(lr){2-3} \cmidrule(lr){4-5}
        ~ & \textbf{CL-7B} & \textbf{CL-13B} & \textbf{CL-7B} & \textbf{CL-13B} \\
        \midrule
        AR (Ref.) & 1.00$\times$ \scriptsize(42\tiny$\pm$1.6) & 1.00$\times$ \scriptsize(25\tiny$\pm$0.7) & 1.00$\times$ \scriptsize(44\tiny$\pm$0.8) & 1.00$\times$ \scriptsize(26\tiny$\pm$0.3) \\
        LADE & 1.45$\times$ \scriptsize(61\tiny$\pm$5.7) & 1.44$\times$ \scriptsize(36\tiny$\pm$4.6) & 1.70$\times$ \scriptsize(75\tiny$\pm$7.5) & 1.65$\times$ \scriptsize(43\tiny$\pm$4.1) \\
        Self-Draft & 1.62$\times$ \scriptsize(68\tiny$\pm$7.4) & 1.68$\times$ \scriptsize(42\tiny$\pm$5.4) & 1.86$\times$ \scriptsize(82\tiny$\pm$6.7) & 1.88$\times$ \scriptsize(49\tiny$\pm$4.3) \\
        \textbf{PTD} & \textbf{1.69$\times$} \scriptsize(\textbf{71}\tiny$\pm$\textbf{7.4}) & \textbf{1.72$\times$} \scriptsize(\textbf{43}\tiny$\pm$\textbf{5.6}) & \textbf{2.05$\times$} \scriptsize(\textbf{90}\tiny$\pm$\textbf{8.9}) & \textbf{2.08$\times$} \scriptsize(\textbf{54}\tiny$\pm$\textbf{5.4}) \\
        \bottomrule
    \end{tabular}
    \caption{Speedup and throughput results on code generation benchmarks. The format follows: $Speedup \times$ \scriptsize(Throughput $\pm$ Std).}
    \label{tab:code-res-full}
\end{table}

Tables~\ref{tab:chat-math-res} and~\ref{tab:code-res-full} summarize the speedup performance. We set the maximum tree depth to 6 and the branching factor to 4. As observed, PTD consistently achieves the highest speedup across all evaluated models and tasks, outperforming state-of-the-art training-free baselines like LADE and Self-Draft.

The advantages of PTD are most pronounced in reasoning-intensive benchmarks, specifically GSM-100 (up to 2.30$\times$) and MBPP-100 (up to 2.08$\times$). Because mathematical and coding tasks follow structured logical paths, PTD’s tree-based evolution effectively maintains high draft coherence while providing broader coverage of the solution space, significantly reducing redundant computations compared to independent branching methods.

Moreover, PTD achieves stable acceleration across various model generations (from LLaMA-2 to Qwen-3) as a training-free, plug-and-play solution. Whether applied to smaller 7B models or the larger Qwen2-32B (reaching a peak speedup of 2.30$\times$), PTD consistently unlocks the model's inherent parallel potential with zero extra learning cost, proving its superior versatility in real-world deployment.

\begin{figure}[h]
    \centering
    \includegraphics[width=\linewidth]{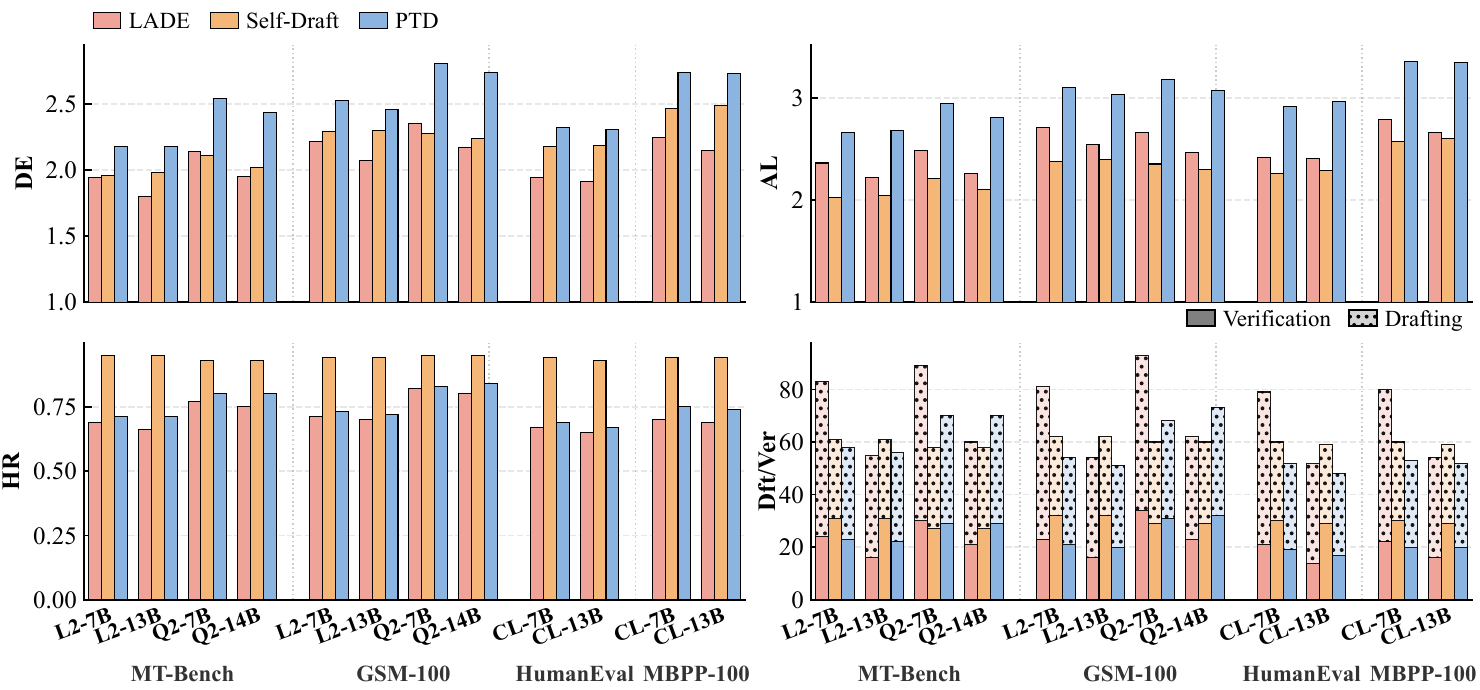}
    \caption{Draft content quality analysis.}
    \label{fig:res-DE}
\end{figure}

Figure~\ref{fig:res-DE} provides a more in-depth analysis of decoding efficiency(DE), hit rate(HR), candidate draft acceptance length(AL), and overhead(Dft/Ver), which reveals the underlying causes of speed differences among the methods. 
On this front, PTD demonstrates a comprehensive advantage. Specifically, while PTD exhibits a slightly lower HR compared to Self-Draft, which primarily because Self-Draft uses specific external data/corpora to boost hit probability, it maintains a substantial lead in AL. 
This indicates that PTD-generated drafts possess superior contextual coherence, allowing the model to accept much longer sequences per verification pass. 

Furthermore, PTD maintains a competitive overhead profile, comparable to Self-Draft and generally lower than LADE. This ensures that the structural complexity of PTD does not translate into significant latency during the forward pass, effectively maximizing the net gain in inference speed.
A detailed runtime breakdown is provided in Appendix~\ref{app:overhead}.

\subsubsection{Draft Tree Analysis}

\begin{figure}[h]
\centering
    \begin{minipage}[t]{0.49\linewidth}
    \centering
        \includegraphics[width=\linewidth]{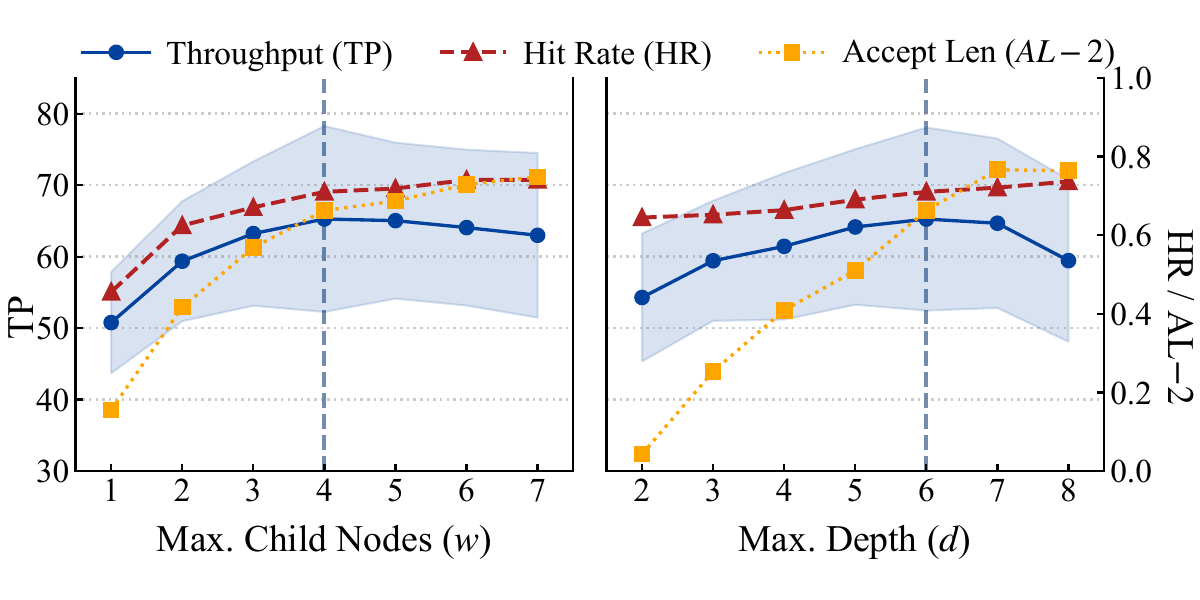}
        \captionof{figure}{Effect of tree width and depth on PTD performance on LLaMA-7B.}
    \label{fig:width-depth}
    \label{fig:7b}
    \end{minipage}
    \begin{minipage}[t]{0.49\linewidth}
    \centering
        \includegraphics[width=\linewidth]{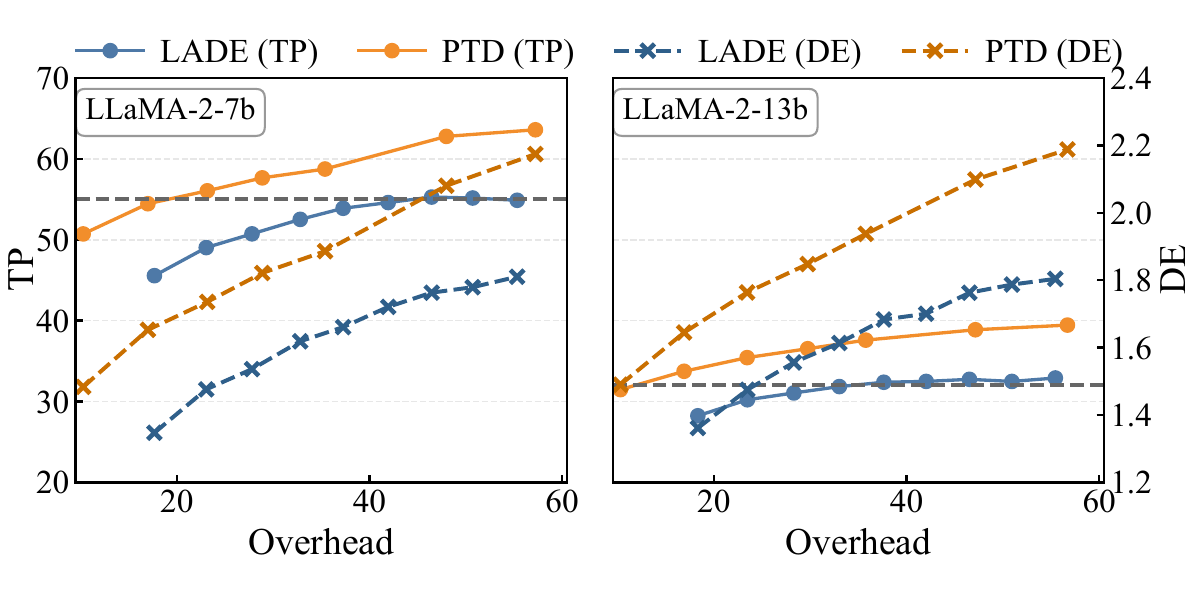}
        \captionof{figure}{Draft efficiency of LADE and PTD on LLaMA-7B and LLaMA-13B.}
    \label{fig:draft_effi}
    \label{fig:13b}
    \end{minipage}
\end{figure}

The performance of PTD is governed by the trade-off between draft quality and computational overhead. Figure~\ref{fig:width-depth} analyzes the effects of the maximum number of child nodes ($w$) and drafting depth ($d$).

\textbf{Impact of Max Child Nodes ($w$).} With depth fixed at 6, increasing $w$ improves both Hit Rate ($HR$) and Accept Length ($AL$) by expanding the search breadth and increasing the chance of covering the intended semantic path. However, throughput follows a rise-then-fall trend and peaks at $w=4$, after which the additional latency of a wider tree outweighs the gains in $HR$ and $AL$.

\textbf{Impact of Max Drafting Depth ($d$).} With $w=4$, $HR$ remains relatively stable as depth increases, suggesting that the initial hit is driven more by branching diversity than depth. In contrast, $AL$ keeps increasing because deeper trees allow longer continuation along a correct path. Throughput peaks at $d=6$, beyond which the benefit of longer accepted sequences is offset by the overhead of verifying deeper trees.

Overall, PTD exhibits strong robustness across a broad range of tree configurations. In practical deployments, $w$ and $d$ can be dynamically adapted based on specific model architectures and hardware resources. In this paper, we utilize a uniform configuration of $w=4$ and $d=6$ to facilitate a consistent discussion across all evaluations.

\subsubsection{Draft efficiency}

Figure~\ref{fig:draft_effi} compares the drafting efficiency of PTD and LADE across different computational overhead levels. PTD consistently achieves higher throughput (TP) and decoding efficiency (DE). On LLaMA-7B, PTD matches LADE's peak throughput with approximately half the overhead; on LLaMA-13B, it requires only about one-third. PTD also provides substantial speedups under low-overhead settings, highlighting its potential for large-scale inference services. Additional evaluations of generation quality under sampling and acceleration under greedy decoding are provided in Appendices~\ref{eval} and~\ref{app:greedy}.

\section{Conclusion}

In this paper, we introduce PTD, a guided and structured inference acceleration method for autoregressive LLMs. This framework achieves comprehensive performance gains across diverse models and benchmarks as a training-free and model-agnostic solution. Future research will explore dense semantic representations, like semantic graphs, to enable more guided and efficient draft generation beyond current tree-based methods. Besides, we also aim to decouple the drafting and decoding processes to enhance system efficiency and scalability.

\bibliography{colm2026_conference}
\bibliographystyle{colm2026_conference}

\appendix
\section{Sample Strategy for the Draft Tree Expansion}
\label{app:sample-strategy}

This section analyzes how the draft-tree expansion strategy affects PTD. Unless otherwise noted, we use LLaMA-2-13B on MT-Bench and compare greedy expansion with top-k and top-p sampling for draft-tree growth. For sampled expansion, we first obtain the top-k or top-p distribution at each node and then sample draft tokens from that distribution to extend the tree.

\begin{figure}[h]
    \centering
    \includegraphics[width=\linewidth]{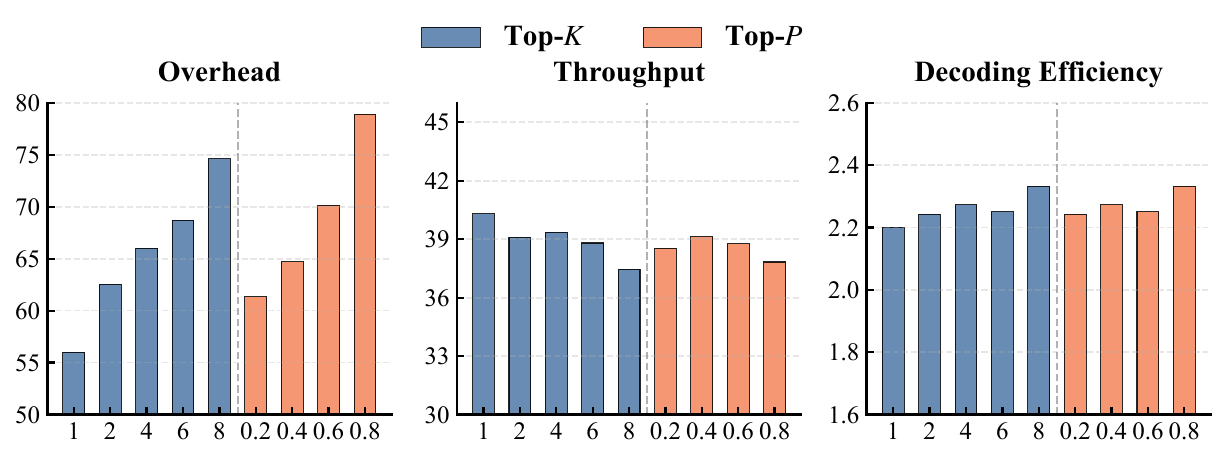}
    \caption{PTD performance under different sampling strategies for draft tree expansion.}
    \label{fig:sample}
\end{figure}

Figure~\ref{fig:sample} shows the resulting trade-off. Compared with greedy expansion, top-k and top-p increase overhead because they introduce greater diversity and uncertainty, which leads to faster tree growth. However, the additional overhead yields little improvement in throughput or decoding efficiency.

\newpage
\section{Impact of Initialization Strategy}
\label{app:init-strategy}

\begin{figure}[ht]
    \centering
    \includegraphics[width=\linewidth]{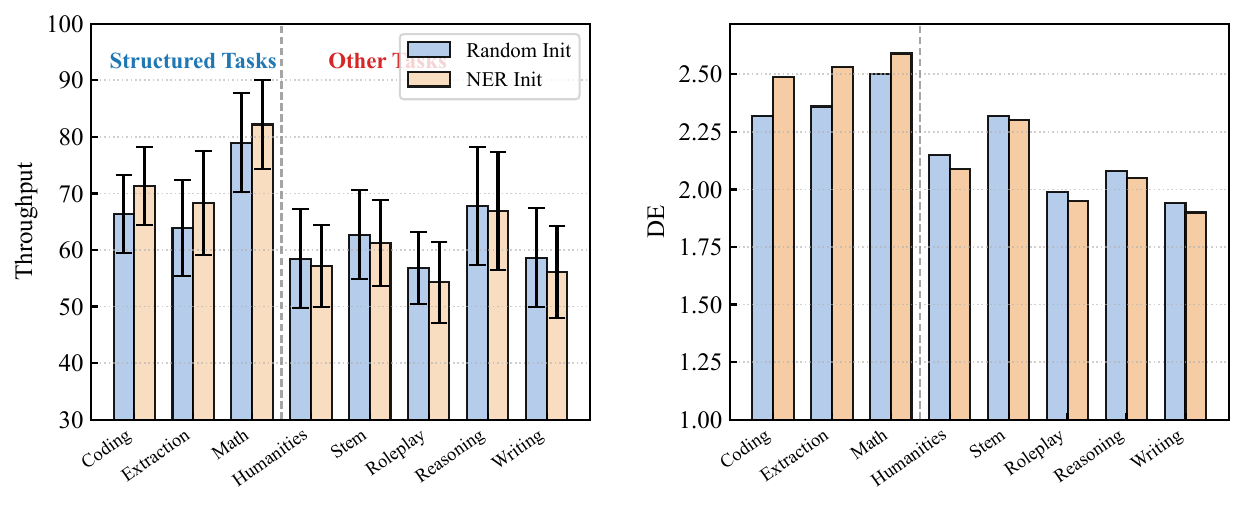}
    \caption{Performance comparison of PTD under random and NER-based initialization across eight MT-Bench tasks. }
    \label{fig:ini-method}
\end{figure}

This section compares random initialization with a Named Entity Recognition (NER)-based initialization strategy for PTD across the eight MT-Bench categories. As shown in Figure~\ref{fig:ini-method}, the effectiveness of the initialization strategy depends on the task type.
For structured domains such as coding, mathematics, and extraction, NER-based initialization performs better because entity-aware seeds better capture task-specific identifiers and core logical elements.
In contrast, the difference between the two methods is small for open-ended tasks such as creative writing. These results suggest that specialized initialization is most useful for structured reasoning and less critical for general-purpose generation.

\newpage
\section{Overhead Analysis}
\label{app:overhead}

\begin{figure}[h]
    \begin{minipage}[ht]{\linewidth}
        \centering
        \includegraphics[width=\linewidth]{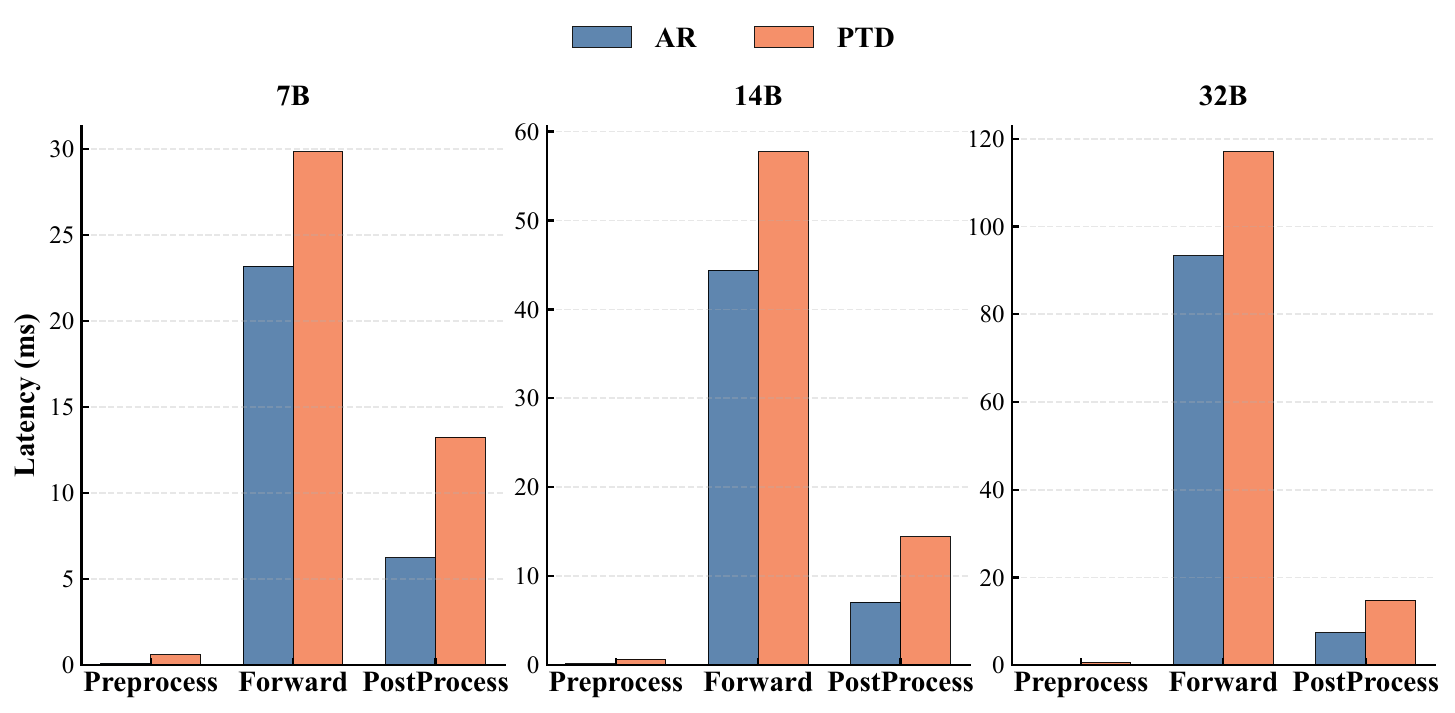}
        \caption{Run time analysis for Qwen models.}
        \label{fig:run_time}
    \end{minipage}
\end{figure}

This section breaks down the runtime overhead of PTD relative to autoregressive decoding on Qwen models. PTD introduces three additional costs: candidate-tree retrieval before the forward pass, parallel drafting and verification during the forward pass, and candidate-pool and draft-tree updates after the forward pass. Figure~\ref{fig:run_time} shows that the dominant overhead comes from the forward pass, where PTD performs additional inference on both the draft tree and the candidate tree, while retrieval and update costs remain negligible.

\newpage
\section{Progressive Tree Drafting Decoding Algorithm}

\label{alg-PTD}

\begin{algorithm}[h]
\caption{Progressive Tree Drafting Decoding Algorithm}
\begin{algorithmic}[1]
\STATE \textbf{Input:} Prompt $\mathbf{X} = [x_1, x_2, ..., x_{t-1}]$; max tree depth $d_{\text{max}}$; initial tree $T^0 = (V^0, E^0)$; max length $N$
\WHILE{True}
    \STATE \{Step 1: Retrieve Candidate Tree\}
    \STATE Retrieve candidate tree $\mathcal{C}_{\mathbf{X}} \gets$ \textsc{RetrieveCandidateTree}($\mathbf{X}$)
    \STATE \{Step 2: Generate Next Token(s) with Structural Guidance\}
    \STATE{$x_t, \mathcal{D}, \mathcal{V} \gets \mathcal{S}\left(P(y_t, \mathbf{y}_T, \mathbf{y}_{\mathcal{C}} \mid \left[\mathbf{X}; T^{i-1}; \mathcal{C}_{\mathbf{X}}\right])\right)$}
    
    \STATE \{Step 3: Expand the Draft Tree\}
    \STATE{$V^i \gets V^{i-1} \cup \{ d_v \mid \forall v \in V^{i-1} \}$}
    \STATE{$E^i \gets E^{i-1} \cup \{ (v, d_v) \mid \forall v \in V^{i-1} \}$}
    
    \IF{$\text{depth}(T^i) > d_{\text{max}}$}
        \STATE{$T^i \gets$ \textsc{StepAndPrune}($T^i$)}
    \ENDIF
    
    \STATE \{Step 4: Merge Subtrees into Candidate Pool\}
    \FOR{each subtree $T'_s$ in $T^i$}        
        \STATE{Update candidate pool by merging trees using $\mathcal{M}$}
    \ENDFOR
    
    \STATE \{Step 5: Obtain Eligible Edges\}
    \IF{Using Greedy Decoding}
        \STATE{$\mathcal{V} \gets \operatorname{argmax}(P_{\mathcal{C}})$}
        \STATE{$\mathcal{E} \gets \{ (n, \mathcal{V}_n) \mid \mathcal{V}_n \in \sigma(n), \forall n \in V^i \}$}
    \ELSIF{Using Sampling Decoding}
        \STATE{$\mathcal{E}, \mathcal{V}_{n_k} \gets$ \textsc{CandidateTreeRecursiveSample}($\mathcal{C}_{\mathbf{X}}$)}
    \ENDIF
    
    \STATE \{Step 6: Append Chosen Path\}
    \STATE{$\mathbf{X}' \gets (n_0, n_1, ..., n_k, \mathcal{V}_{n_k})$ s.t. $\forall i < k, (n_i, n_{i+1}) \in \mathcal{E}$}
    \STATE{Append $\mathbf{X}'$ to $\mathbf{X}$}
    
    \IF{$|\mathbf{X}| > N$}
        \STATE \textbf{break}
    \ENDIF
    
    \STATE $i \leftarrow i + 1$
\ENDWHILE
\STATE \textbf{Output:} Generated sequence $\mathbf{X}$
\end{algorithmic}
\end{algorithm}

\newpage
\section{Candidate Tree Recursive Sampling Algorithm}
\label{alg-sample}
\begin{algorithm}
\caption{Candidate Tree Recursive Sampling}
\begin{algorithmic}[1]
\STATE \textbf{Input:} A node $v$
\STATE \textbf{Output:} Obtain global eligible edges $\mathcal{E}$
\STATE $C \gets \sigma(v)$ \COMMENT{Children of $v$}
\WHILE{$C$ is not empty}
    \FORALL{$n \in C$}
        \STATE Sample $r \sim \mathcal{U}(0, 1)$
        \IF{$r < P(n)$}
            \STATE Append $(v,n)$ to $\mathcal{E}$
            \STATE \textbf{call} $\mathcal{V}_{n_k} \gets$ \textsc{Traversal}$(n)$
            \RETURN $n_k$
        \ELSE
            \STATE $P[n] \gets 0$
            \STATE Renormalize $P$ over remaining nodes in $C$
        \ENDIF
    \ENDFOR
\ENDWHILE
\STATE \COMMENT{If no child selected, sampling based on current node distribution}
\RETURN  $\mathcal{S}\left(P(v)\right)$ 
\end{algorithmic}
\end{algorithm}

\newpage

\section{Proof of Distributional Consistency of the Candidate Tree Recursive Sampling Algorithm}
\label{proof}

We aim to prove that the sampling algorithm described in Appendix~\ref{alg-PTD} selects each candidate node $n_i$ with probability equal to its original probability $P_i$.

\paragraph{Sampling Procedure.}
Given a set of candidate nodes $\{n_1, n_2, \dots, n_k\}$ and associated probabilities $P_i$, the algorithm iteratively samples a random variable $r \sim \mathcal{U}(0, 1)$ and accepts the first node $n_i$ such that $r < P_i$ (after re-normalization, if any earlier nodes have been rejected). If $n_i$ is not accepted, its probability is set to 0, and the remaining probabilities are re-normalized.

\paragraph{Objective.}
Let $\mathcal{A}_i$ denote the event that node $n_i$ is selected. We aim to prove:
\[
\mathbb{P}(\mathcal{A}_i) = P_i, \quad \forall i \in \{1, 2, \dots, k\}.
\]

\paragraph{Base Case ($i = 1$).}
Node $n_1$ is the first candidate considered. Since no re-normalization has occurred yet, its acceptance probability is:
\[
\mathbb{P}(\mathcal{A}_1) = \mathbb{P}(r < P_1) = P_1.
\]

\paragraph{Inductive Step.}
Suppose that for each $j < i$, the probability of selecting node $n_j$ is exactly $P_j$, and the algorithm correctly rejects $n_1$ through $n_{i-1}$ with total probability $R_{i-1} = \sum_{j=1}^{i-1} P_j$.

After rejecting $n_1, \dots, n_{i-1}$, the remaining unnormalized probability is:
\[
S_{i-1} = 1 - \sum_{j=1}^{i-1} P_j.
\]

The normalized probability of $n_i$ in this residual distribution becomes:
\[
\hat{P}_i = \frac{P_i}{S_{i-1}}.
\]

The probability of reaching $n_i$ without accepting any of the previous $i-1$ nodes is:
\[
\mathbb{P}(\text{reaching } n_i) = \prod_{j=1}^{i-1} (1 - \hat{P}_j).
\]

However, since:
\[
\prod_{j=1}^{i-1} (1 - \hat{P}_j) = \prod_{j=1}^{i-1} \left(1 - \frac{P_j}{S_{j-1}} \right) = \frac{S_1}{S_0} \cdot \frac{S_2}{S_1} \cdots \frac{S_{i-1}}{S_{i-2}} = \frac{S_{i-1}}{S_0} = S_{i-1},
\]
and $S_0 = 1$, this implies:
\[
\mathbb{P}(\text{reaching } n_i) = S_{i-1}.
\]

Therefore, the total probability of accepting $n_i$ is:
\[
\mathbb{P}(\mathcal{A}_i) = \mathbb{P}(\text{reaching } n_i) \cdot \hat{P}_i = S_{i-1} \cdot \frac{P_i}{S_{i-1}} = P_i.
\]

\paragraph{Conclusion.}
By induction, for every $i \in \{1, \dots, k\}$, the probability of node $n_i$ being selected is exactly $P_i$. Hence, the sampling algorithm yields a sample from the original distribution $P$:
\[
\mathbb{P}(\mathcal{A}_i) = P_i \quad \forall i.
\]
This proves that the sequential rejection-normalization sampling procedure preserves the target distribution.

\newpage
\section{Generation Quality Evaluation: A Comparison Between PTD and Autoregressive Decoding under the Sampling Strategy}

\label{eval}
\setlength{\tabcolsep}{1.5pt}
This section compares the generated content of PTD and standard autoregressive decoding under the sampling setting. We evaluate MT-Bench, GSM-100, HumanEval, and MBPP-100 with LLaMA-2 (L), Qwen-2.5 (Q), and CodeLLaMA (CL) models using Rouge-1, Rouge-2, Rouge-L, and BLEU.
\begin{table}[!h]
    \centering
    \begin{tabular}{llcccc}
    \toprule
        \textbf{Benchmark} & \textbf{Model} & \textbf{Rouge-1} & \textbf{Rouge-2} & \textbf{Rouge-L} & \textbf{BLEU} \\ 
        \midrule
        \multirow{5}{*}{MT-Bench} & L-7B & 50 & 32 & 34 & 17 \\ 
        ~ & L-13B & 51 & 34 & 36 & 19 \\ 
        ~ & Q-7B & 42 & 20 & 24 & 21 \\ 
        ~ & Q-14B & 48 & 22 & 24 & 18 \\ 
        ~ & Q-32B & 48 & 24 & 26 & 22 \\ 
        \midrule
        \multirow{5}{*}{GSM-100} & L-7B & 68 & 53 & 55 & 39 \\ 
        ~ & L-13B & 65 & 50 & 53 & 36 \\ 
        ~ & Q-7B & 49 & 31 & 34 & 26 \\ 
        ~ & Q-14B & 52 & 29 & 31 & 28 \\ 
        ~ & Q-32B & 58 & 40 & 41 & 38 \\ 
        \midrule
        \multirow{2}{*}{HumanEval} & CL-7B & 48 & 38 & 40 & 26 \\ 
        ~ & CL-13B & 48 & 40 & 43 & 21 \\ 
        \midrule
        \multirow{2}{*}{MBPP-100} & CL-7B & 82 & 77 & 80 & 77 \\ 
        ~ & CL-13B & 82 & 78 & 80 & 76 \\ 
        \bottomrule
    \end{tabular}
    \caption{Comparison of generated content between PTD and autoregressive decoding under the sampling strategy. L, Q, and CL denote LLaMA-2, Qwen-2.5, and CodeLLaMA, respectively.}
\end{table}

\newpage
\section{Acceleration Performance of Greedy Decoding Strategy}
\label{app:greedy}

This section reports acceleration results under greedy decoding on MT-Bench, GSM-100, HumanEval, and MBPP-100 using LLaMA-2 (L), Qwen-2.5 (Q), and CodeLLaMA (CL) models.

\begin{table*}[htbp]
    \centering
    \begin{tabular}{lllllllll}
        \toprule
        \multirow{2}{*}{\textbf{Benchmark}} & \multirow{2}{*}{\textbf{Model}} & \textbf{AR} & \multicolumn{2}{c}{\textbf{LADE}} & \multicolumn{2}{c}{\textbf{Self-Draft}} & \multicolumn{2}{c}{\textbf{PTD}} \\ 
        \cmidrule(lr){4-5} \cmidrule(lr){6-7} \cmidrule(lr){8-9}
        ~ & ~ & TP\tiny(Std) & TP\tiny(Std) & Imp. & TP\tiny(Std) & Imp. & TP\tiny(Std) & Imp. \\ 
        \midrule
        \multirow{5}{*}{MT-Bench} & L-7B & 40\text{\tiny$\pm$4.1} & 59\text{\tiny$\pm$9.4} & 47\% & 62\text{\tiny$\pm$11.4} & 56\% & \textbf{67}\text{\tiny$\pm$10.8} & \textbf{68\%} \\ 
        ~ & L-13B & 24\text{\tiny$\pm$1.7} & 34\text{\tiny$\pm$4.8} & 41\% & 37\text{\tiny$\pm$6.7} & 54\% & \textbf{40}\text{\tiny$\pm$6.4} & \textbf{67\%} \\ 
        ~ & Q-7B & 36\text{\tiny$\pm$4.4} & 59\text{\tiny$\pm$12.2} & 65\% & 55\text{\tiny$\pm$13.4} & 52\% & \textbf{70}\text{\tiny$\pm$20.0} & \textbf{93\%} \\ 
        ~ & Q-14B & 20\text{\tiny$\pm$2.0} & 31\text{\tiny$\pm$5.3} & 57\% & 31\text{\tiny$\pm$6.3} & 56\% & \textbf{36}\text{\tiny$\pm$6.8} & \textbf{81\%} \\ 
        ~ & Q-32B & 10\text{\tiny$\pm$0.6} & 16\text{\tiny$\pm$2.7} & 57\% & 16\text{\tiny$\pm$3.3} & 62\% & \textbf{19}\text{\tiny$\pm$3.6} & \textbf{88\%} \\ 
        \midrule
        \multirow{5}{*}{GSM-100} & L-7B & 44\text{\tiny$\pm$1.0} & 74\text{\tiny$\pm$5.9} & 66\% & 75\text{\tiny$\pm$6.7} & 68\% & \textbf{85}\text{\tiny$\pm$7.2} & \textbf{91\%} \\ 
        ~ & L-13B & 26\text{\tiny$\pm$0.4} & 41\text{\tiny$\pm$3.3} & 58\% & 44\text{\tiny$\pm$4.6} & 67\% & \textbf{49}\text{\tiny$\pm$4.3} & \textbf{89\%} \\ 
        ~ & Q-7B & 40\text{\tiny$\pm$2.1} & 72\text{\tiny$\pm$8.2} & 80\% & 65\text{\tiny$\pm$8.8} & 62\% & \textbf{86}\text{\tiny$\pm$16.4} & \textbf{116\%} \\ 
        ~ & Q-14B & 22\text{\tiny$\pm$0.6} & 37\text{\tiny$\pm$3.7} & 67\% & 37\text{\tiny$\pm$4.6} & 69\% & \textbf{44}\text{\tiny$\pm$5.2} & \textbf{99\%} \\ 
        ~ & Q-32B & 11\text{\tiny$\pm$0.2} & 19\text{\tiny$\pm$1.6} & 81\% & 19\text{\tiny$\pm$1.2} & 82\% & \textbf{24}\text{\tiny$\pm$2.2} & \textbf{125\%} \\
        \midrule
        \multirow{2}{*}{HumanEval} & CL-7B & 43\text{\tiny$\pm$1.7} & 62\text{\tiny$\pm$6.8} & 45\% & 62\text{\tiny$\pm$7.6} & 45\% & \textbf{74}\text{\tiny$\pm$8.5} & \textbf{74\%} \\ 
        ~ & CL-13B & 25\text{\tiny$\pm$0.7} & 37\text{\tiny$\pm$4.5} & 45\% & 39\text{\tiny$\pm$5.3} & 55\% & \textbf{44}\text{\tiny$\pm$5.8} & \textbf{74\%} \\ 
        \midrule
        \multirow{2}{*}{MBPP-100} & CL-7B & 45\text{\tiny$\pm$0.8} & 77\text{\tiny$\pm$6.4} & 71\% & 73\text{\tiny$\pm$7.2} & 62\% & \textbf{93}\text{\tiny$\pm$9.9} & \textbf{108\%} \\ 
        ~ & CL-13B & 26\text{\tiny$\pm$0.3} & 43\text{\tiny$\pm$4.1} & 64\% & 48\text{\tiny$\pm$4.5} & 82\% & \textbf{55}\text{\tiny$\pm$5.4} & \textbf{107\%} \\ 
        \bottomrule
    \end{tabular}
    \caption{Throughput and relative improvement (Imp.) under greedy decoding for autoregressive decoding (AR), LADE, Self-Draft, and PTD. L, Q, and CL denote LLaMA-2, Qwen-2.5, and CodeLLaMA, respectively.}
    \label{main-tp-res}
\end{table*}

\setlength{\tabcolsep}{2pt}
\begin{table*}[htbp]
    \centering
    \begin{tabular}{llllllllllllll}
    \toprule
        \multirow{2}{*}{\textbf{Benchmark}} & \multirow{2}{*}{\textbf{Model}} & \multicolumn{4}{c}{\textbf{LADE}} & \multicolumn{4}{c}{\textbf{Self-Draft}} & \multicolumn{4}{c}{\textbf{PTD}} \\ 
        \cmidrule(lr){3-6} \cmidrule(lr){7-10} \cmidrule(lr){11-14}
        ~ & ~ & DE & HR & AL & Dft/Ver & DE & HR & AL & Dft/Ver & DE & HR & AL & Dft/Ver \\ 
        \midrule
        \multirow{5}{*}{MT-Bench} & L-7B & 1.95 & 0.69 & 2.39 & 59/23 & 1.96 & 0.95 & 2.02 & 30/30 & \textbf{2.23} & 0.71 & \textbf{2.74} & 35/23 \\ 
        ~ & L-13B & 1.83 & 0.67 & 2.26 & 39/17 & 1.96 & 0.95 & 2.02 & 30/30 & \textbf{2.20} & 0.71 & \textbf{2.70} & 34/22 \\ 
        ~ & Q-7B & 2.20 & 0.78 & 2.55 & 59/31 & 2.03 & 0.92 & 2.12 & 31/26 & \textbf{2.58} & 0.80 & \textbf{2.99} & 40/29 \\ 
        ~ & Q-14B & 2.01 & 0.76 & 2.31 & 39/21 & 1.97 & 0.92 & 2.05 & 31/26 & \textbf{2.40} & 0.80 & \textbf{2.76} & 42/29 \\ 
        ~ & Q-32B & 1.87 & 0.72 & 2.21 & 27/15 & 2.02 & 0.92 & 2.11 & 31/25 & \textbf{2.43} & 0.77 & \textbf{2.86} & 37/27 \\ 
        \midrule
        \multirow{5}{*}{GSM-100} & L-7B & 2.23 & 0.72 & 2.72 & 58/22 & 2.29 & 0.94 & 2.38 & 30/32 & \textbf{2.52} & 0.73 & \textbf{3.09} & 32/21 \\ 
        ~ & L-13B & 2.06 & 0.70 & 2.53 & 38/16 & 2.29 & 0.94 & 2.38 & 30/32 & \textbf{2.48} & 0.72 & \textbf{3.05} & 31/20 \\ 
        ~ & Q-7B & 2.44 & 0.83 & 2.75 & 59/35 & 2.25 & 0.95 & 2.32 & 31/28 & \textbf{2.90} & 0.84 & \textbf{3.26} & 37/31 \\ 
        ~ & Q-14B & 2.16 & 0.80 & 2.45 & 39/23 & 2.19 & 0.95 & 2.25 & 31/28 & \textbf{2.68} & 0.84 & \textbf{3.00} & 41/32 \\ 
        ~ & Q-32B & 2.16 & 0.80 & 2.46 & 27/17 & 2.34 & 0.96 & 2.40 & 31/29 & \textbf{2.91} & 0.84 & \textbf{3.28} & 35/30 \\ 
        \midrule
        \multirow{2}{*}{HumanEval} & CL-7B & 1.96 & 0.67 & 2.44 & 58/20 & 2.15 & 0.94 & 2.24 & 30/30 & \textbf{2.35} & 0.69 & \textbf{2.97} & 33/19 \\ 
        ~ & CL-13B & 1.95 & 0.66 & 2.45 & 38/15 & 2.23 & 0.93 & 2.33 & 30/29 & \textbf{2.35} & 0.68 & \textbf{3.00} & 31/17 \\ 
        \midrule
        \multirow{2}{*}{MBPP-100} & CL-7B & 2.29 & 0.71 & 2.82 & 58/23 & 2.47 & 0.94 & 2.57 & 30/30 & \textbf{2.75} & 0.74 & \textbf{3.36} & 33/20 \\ 
        ~ & CL-13B & 2.13 & 0.69 & 2.63 & 38/16 & 2.48 & 0.94 & 2.59 & 30/30 & \textbf{2.72} & 0.74 & \textbf{3.34} & 32/19 \\ 
        \bottomrule
    \end{tabular}
    \caption{Decoding efficiency (DE), hit rate (HR), Accept Length (AL), and overheads (Dft/Ver) of PTD, LADE, and Self-Draft under greedy decoding.}

    \label{res-DE}
\end{table*}

\end{document}